# Image Based Camera Localization: an Overview

Yihong Wu, Fulin Tang, Heping Li

National Laboratory of Pattern Recognition, Institute of Automation, Chinese Academy of Sciences

University of Chinese Academy of Sciences

yhwu@nlpr.ia.ac.cn

## Abstract

Recently, virtual reality, augmented reality, robotics, autonomous driving et al attract much attention of both academic and industrial community, in which image based camera localization is a key task. However, there has not been a complete review on image-based camera localization. It is urgent to map this topic to help people enter the field quickly. In this paper, an overview of image based camera localization is presented. A new and complete kind of classifications for image based camera localization is provided and the related techniques are introduced. Trends for the future development are also discussed. It will be useful to not only researchers but also engineers and other people interested.

**Keywords:** PnP problem, SLAM, Camera localization, Camera pose determination

## 1. Introduction

Recently, virtual reality, augmented reality, robotics, autonomous driving et al attract much attention of both academic and industrial community, in which image based camera localization is a key task. It is urgent to give an overview of image based camera localization.

The sensors of image based camera localization are cameras. There occur many 3D cameras recently. This paper only considers 2D cameras. The usually used tool for localization is GPS outdoors. However, GPS cannot be used indoors. There are many indoor localization tools including Lidar, UWB, WiFi AP et al, among which using cameras to localization is the most flexible and low cost one. Autonomous localization and navigation is necessary for a moving robot. To augment reality on images, the camera pose or localization are needed. To view virtual environments, the corresponding viewing angle is necessarily computed. Furthermore, cameras are ubiquitous and people carry with their mobile phones that have cameras every day. Image based camera localization has great and wide applications.

The used image features of image based camera localization are points, lines, conics, spheres and angles. The widely used ones are from points. This paper pays more attention to points.

Image based camera localization is a large topic, we try to cover the related works and give a complete kind of classifications for image based camera localization. But it is still possible to miss out some works in a limited space of a paper. Also we cannot give deep criticism for each cited paper due to space limit for a large topic. Further deep reviews on some hot aspect of image based camera localization will be given in the future or people interested go to read already existing surveys. There have been some excellent reviews on some aspects of the image based camera localization. The most recent ones are as follows. [Khan and Adnan 2016] gave an overview of ego motion estimation, where ego motion requires that time intervals between two continuous images is small enough. [Cadena et al 2016] surveyed the current state of SLAM and considered future directions, in which they reviewed related works including robustness and scalability in long-term mapping, metric and semantic representations for mapping, theoretical performance guarantees, active SLAM and exploration. [Yunes et al 2017] specially reviewed the keyframe-based monocular SLAM. [Piasco et al 2018] gave a survey on visual-based localization from heterogeneous data, where only known environment is considered.

Different from them, this paper is unique in that it first maps the whole image-based camera localization and gives a complete kind of classifications for the topic. Section 2 presents the overview of image-based camera localization and map it by a tree figure. Section 3 introduces each aspect of the classifications. Section 4 makes discussions and analyzes trends of the future developments.

## 2. Overview

What is the image based camera localization? It is to compute camera poses under some world coordinate system from images or videos captured by the cameras. According to that environments are prior or not, image based camera localization has two categories: the one with known environments and the other one with unknown environments.

Let $n$ be the number of used points. The one with known environments consists of the methods with $3 \leq n < 6$ and the methods with $n \geq 6$. These are the PnP problem studies. In general, the problems with $3 \leq n < 6$ are nonlinear and the problems with $n \geq 6$ are linear.

The one with unknown environments consists of the methods with online and real time environment mapping and the methods without online and real time environment mapping. The former is the commonly known Simultaneous Localization and Mapping (SLAM) and the latter is the middle procedure of the commonly known structure from motion (SFM). According to different map generations, SLAM has four parts: geometric metric SLAM, learning SLAM, topological SLAM, and marker SLAM. Learning SLAM is a new research direction recently. We think it is a single category different from geometric metric SLAM and topological SLAM. Learning SLAM can obtain camera pose and 3D map but need prior dataset to train a network. The performance of learning SLAM depends on the used dataset greatly and has low generalization ability. Therefore, learning SLAM is not as flexible as geometric metric SLAM and its obtained 3D map outside the used dataset is not as accurate as geometric metric SLAM at the most. But simultaneously, learning SLAM has 3D map other than topology representations. Marker SLAM is to compute camera poses from some known structured markers without knowing the whole environments. Geometric metric SLAM consists of monocular SLAM, multiocular SLAM, and multi-kind sensors SLAM. Also, geometric metric SLAM can be divided as filter-based SLAM and keyframe-based SLAM. Keyframe-based SLAM can be further divided into feature-based SLAM and direct SLAM. Multi-kind sensors SLAM can be divided into loosely-coupled SLAM and closely-coupled SLAM. These classifications of image based camera localization methods are visualized as a tree by logic in Figure 1, where the current hot topics are indicated by bold boarders. We think the current hot topics are camera localization from large data, learning SLAM, keyframe-based SLAM, and multi-kind sensors SLAM.

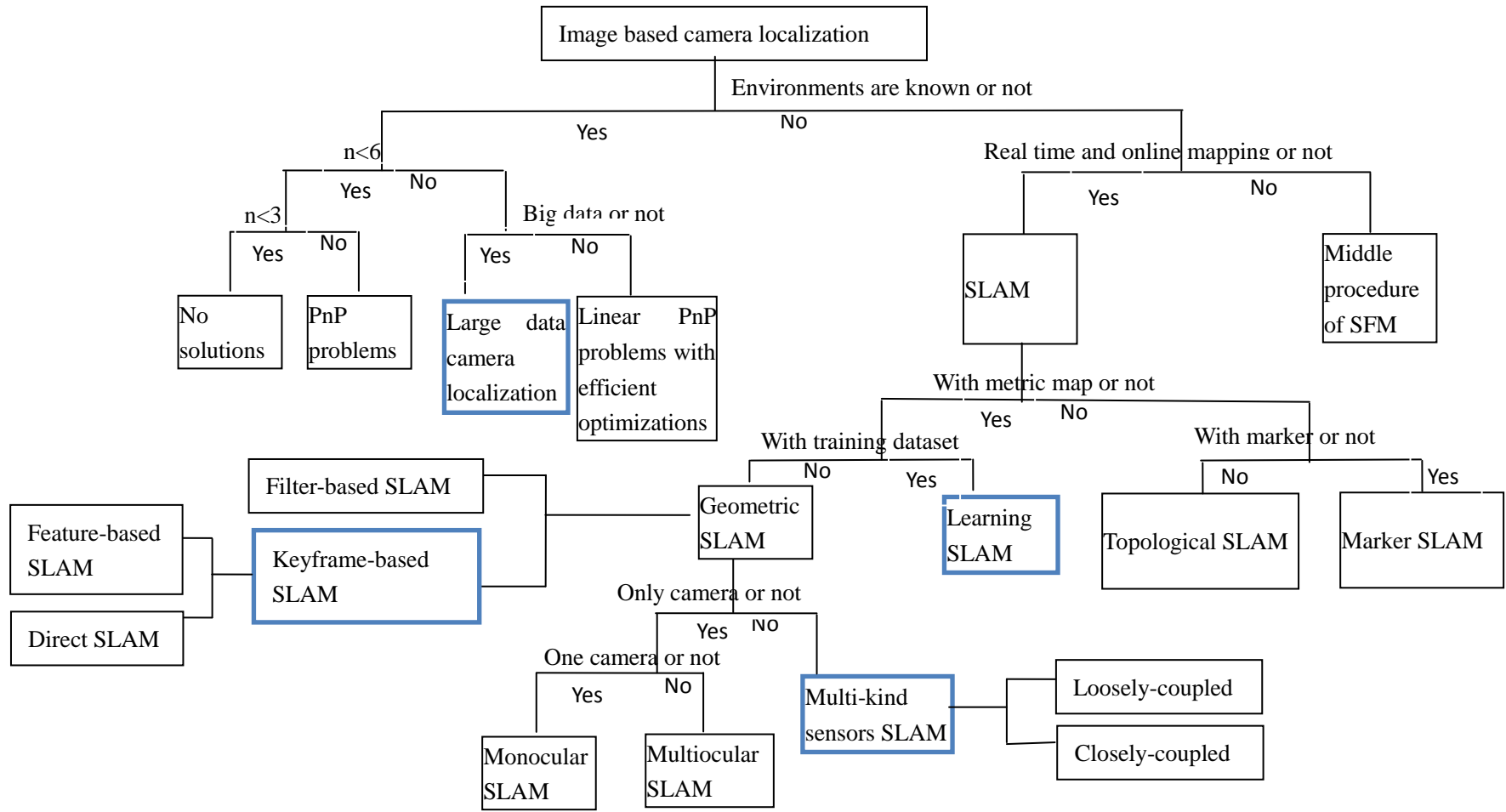

Figure 1. An overview of image based camera localization

## 3. Reviews on image based camera localization

### 3.1 In known environments

Camera pose determinations from known 3D space points are called perspective-n-point problem, namely, PnP problem. When n=1,2, there are no solutions for the PnP problems because they are under constraints. When $n \geq 6$, the PnP problems are linear. When $n = 3,4,5$, original equations of the PnP problems are usually nonlinear. The PnP problem dated from 1841 and 1903. [Grunert 1841] and [Finsterwalder and Scheufele 1903] gave the conclusion that P3P problem has at most 4 solutions and P4P problem has an unique solution in general. The PnP problem is also the key relocalization for SLAM.

### 3.1.1 The PnP problems with $n = 3,4,5$

The works of the PnP problems with $n = 3,4,5$ pay attention to two aspects. One aspect studies the solution numbers or multi solution geometric configuration of the nonlinear problems. The other aspect studies the eliminations or other solving methods for camera poses.

The works of the first aspect are as follows. [Grunert 1841, Finsterwalder and Scheufele 1903] pointed out that P3P has up to four solutions and P4P has a unique solution. [Fischler and Bolles 1981] studied P3P for RANSAC of PnP and found that four solutions of P3P are attainable. [Wolfe et al 1991] gave that P3P has two solutions at the most time as well as the two solution determinations and gave the geometric meanings of that P3P has two, three, four solutions. Hu and Wu 2002 defined distance-based and transformation-based P4P problems. They found the two defined P4P problems are not equivalent, of which transformation-based has up to 4 solutions and distance-based has up to 5 solutions. [Zhang and Hu 2005] provided sufficient and necessary conditions of that P3P has four solutions. [Wu and Hu 2006] proved that distance-based is equivalent to rotation-transformation-based for P3P and distance-based is equivalent to orthogonal-transformation-based for P4P/P5P. Also, they showed that for any three non-collinear points, a location of optical center can always be found such that the P3P problem formed by these three control points and the optical center can have 4 solutions, its upper bound. Additionally a geometric way is provided to construct these 4 solutions. [Vynnycky and Kanev 2015] studied the multi solution probabilities of equilateral P3P problem.

The works of the second aspect of the PnP problems with $n = 3,4,5$ are as follows. [Horaud et al 1989] gave an elimination method for P4P problem to obtain a unitary quartic equation. [Haralick et al 1991] reviewed six methods of P3P that are [Grunert 1841]，[Finsterwalder 1903]，[Merritt 1949]，[Fischler and Bolles 1981], [Hung et al. 1985], [Linnainmaa et al. 1988], and [Grafarend et al. 1989]. [Dementhon and Davis 1992] presented the solution of P3P problem by inquiry table of qusi-perspective imaging. [Quan and Lan 1999] linearly solved the P4P and P5P problem. [Gao et al 2003] took Wu's elimination method to obtain the complete solutions of P3P problem. [Wu and Hu 2006] introduced a depth-ratio based approach to represent the solutions of the whole PnP problem. [Josephson and Byrod 2009] took Grobner bases method to solve P4P problem for the radial distortion camera with unknown focal length. [Hesch et al 2011] studied nonlinear square solutions of PnP with n>=3. [Kneip et al 2011] directly solved the rotation and translation solutions of P3P problem. [Kneip et al 2014] presented a unified PnP solution that can deal with generalized cameras and multi-solutions with global optimizations and linear complexity. [Kuang et al 2013] studied the PnP problem with unknown focal length using points and lines. [Bujnak et al 2013] studied the PnP problem with unknown focal length from radial distortion images. [Ventura et al 2014b] gave a minimal solution to the generalized pose-and-scale problem. [Zheng et al 2014] introduced the angle constraint and derive a compact bivariate polynomial equation for each P3P, then proposed a general method for the PnP problem with unknown focal lendth by iterations. Later, [Zheng et al 2016] improved their work of 2014 with no need point order and iterations. [Wu 2015] studied the PnP solutions with unknown focal length and n=3.5. [Albl et al 2015] studied the pose solution of a rolling shutter camera and improved the result later in 2016.

### 3.1.2 The PnP problems with $n \geq 6$

At the time, the PnP problems are linear and their studies have two aspects. One aspect studies efficient optimizations for the camera poses from smaller number of points. The other aspect studies fast camera localization from large data.

The works of the first aspect are as follows. [Lu et al 2000] gave a global convergence algorithm using collinear points. [Schweighofer and Pinz 2006] studied the multi-solutions of a planar target. [Wu, Hu, and Li 2008] presented invariant relationships between scenes and images, then a robust RANSAC PNP using the invariants. [Lepetit et al 2009] provided an accurate O(n) solution to the PnP problem, called EPnP which is widely used today. The pose problem of a rolling shutter camera is studied by [Hedborg et al 2012] with bundle adjustments. The similar problem is also studied by [Oth et al 2013] using B spline covariance matrix. [Zheng et al 2013] used quaternion and Grobner bases to provide a global optimized solution of the PnP problem. A very fast solution to the PnP Problem with algebraic outlier rejection is given by [Ferraz et al 2014]. [Svarm et al 2014] studied an accurate localization and pose estimation for large 3D models considering gravitational direction. [Ozyesil and Singer 2015] gave a robust camera location estimation by convex programming. [Brachmann et al 2016] gave an uncertainty-driven 6D pose estimation of objects and scenes from a single RGB image. [Feng-Tian-Zhang-Sun 2016] proposed a hand-eye calibration free strategy to actively relocate camera into the same 6D pose by sequentially correcting 3D relative rotation and translation. [Nakano 2016] solved three kinds of PnP problem by Grobner method: PnP problem for calibrated camera, PnPf problem for cameras with unkown focal length, PnPfr problem for cameras with unknown focal length and unknown radial distortions.

The works of the second aspect that pay attention to fast camera localization from large data are as follows. [Arth et al 2009, Arth et al 2011] presented real time camera localizations on mobile phones. [Sattler et al 2011] derived a direct matching framework based on visual vocabulary quantization and a prioritized correspondence search with known large scale 3D models of urban scenes. Later, they improved the method by active correspondence search in [Sattler et al 2012]. [Li et al 2010] devised an adaptive, prioritized algorithm for matching a representative set of SIFT features covering a large scene to a query image for efficient localization. Later [Li et al 2012] gave a full 6-DOF-plus-intrincs camera pose with respect to a large geo-registered 3D point cloud. [Lei et al 2014] studied an efficient camera localization from street views by PCA-based point grouping. [Bansal and Daniilidis 2014] proposed a purely geometric correspondence-free approach to urban geo-localization using 3D point-ray features extracted from the Digital Elevation Map of an urban environment. [Kendall et al 2015] presented a robust and real-time monocular six degree of freedom relocalization system by training a convolutional neural network to regress the 6-DOF camera pose from a single RGB image in an end-to-end manner. [Wang et al 2015] proposed a novel approach to localization in very large indoor spaces that takes a single image and a floor plan of the environment as input. [Zeisl et al 2015] proposed a voting-based pose estimation strategy that exhibits O(n) complexity in the number of matches and thus facilitates to consider much more matches. [Lu et al 2015] used a 3D model reconstructed by a short video as the query to realize 3D-to-3D localization under a multi-task point retrieval framework. [Valentin et al 2015] trained a regression forest to predict mixtures of anisotropic 3D Gaussians and show how the predicted uncertainties can be taken into account for continuous pose optimization. [Straub et al 2013] proposed a relocalization system that enables realtime, 6D pose recovery for wide baselines by using binary feature descriptors and nearest-neighbor search of Locality Sensitive Hashing. [Feng-Fan-Wu 2016] gave a fast localization in large-scale environments by using supervised indexing of binary features, where randomized trees are constructed in a supervised training process by exploiting the label information derived from that multiple features correspond to a common 3D point. [Ventura et al 2012] proposed a system of arbitrary wide-area environments for real-time tracking with a handheld device. The combination of a keyframe-based monocular SLAM system and a global localization method is presented by [Ventura et al 2014a]. A book of large-scale visual geo-localization was published by [Zamir et al. 2016]. [Liu et al 2017] gave an efficient global 2D-3D matching for camera localization in a large-scale 3D map. [Campbell 2017] presented a method of globally-optimal inlier set maximisation for simultaneous camera pose and feature correspondence. A real-time SLAM relocalization with on-line learning of binary feature indexing was proposed by [Feng et al 2017]. [Wu et al 2017] proposed convolutional neural networks for camera relocalization. [Kendall and Cipolla 2017] explored a number of novel loss functions for learning camera pose which are based on geometry and scene reprojection error. [Qin et al 2018] developed a method of relocalization for monocualr visual-inertial SLAM. [Piasco et al 2018] gave a survey on visual-based localization from heterogeneous data.

From above the works of known environments, we see that fast camera localization from large data attract more and more attentions. This is because there are many applications for camera localization from large data, for example location based service, relocalization of SLAM for all kinds of robots, and AR navigations.

## 3.2 In anonymous environments

The unknown environments can be reconstructed from videos real time and online. Simultaneously, the camera poses are computed real time and online. This is the commonly known SLAM technologies. If the unknown environments are reconstructed from multi view images without requiring speed and online, this is the known structure from motion (SFM), in which camera pose solving is a middle procedure and not the final aim so we only mention some works of SFM and don't introduce it deeply in the following. Works of SLAM will be introduced in details.

### 3.2.1 SLAM

SLAM dated from 1986 in the paper of R.C. Smith and Peter Cheeseman: "On the representation and estimation of spatial uncertainty", published in the International Journal of Robotics Research. In 1995, the acronym 'SLAM' was then coined in the paper of H. Durrant-Whyte, D. Rye, and E. Nebot: "Localisation of automatic guided vehicles", The 7th International Symposium on Robotics Research. According to different map generations, the studies of SLAM has four categories: Geometric metric SLAM, learning SLAM, topological SLAM, and marker SLAM. Due to the accurate computations, geometric metric SLAM attract more and more attentions. Learning SLAM is a new topic recently due to the deep learning development. Studies of pure topological SLAM are less and less. Marker SLAM is more accurate and stable. There is a paper on review of the recent advances of SLAM by Cadena et al in 2016 covering a broad set of topics including robustness and scalability in long-term mapping, metric and semantic representations for mapping, theoretical performance guarantees, active SLAM and exploration. In the following, we introduce geometric metric SLAM, learning SLAM, topological SLAM, and marker SLAM respectively.

**A. Geometric Metric SLAM**

Geometric Metric SLAM is to compute 3D maps with accurate mathematical equations. According to using different sensors, Geometric metric SLAM consists of monocular SLAM, multiocular SLAM, and multi-kind sensors SLAM. According to using different techniques, geometric metric SLAM consists of filter-based SLAM, keyframe-based SLAM, and grid-based SLAM. Recently, there is a paper on review of the keyframe-based monocular SLAM and gave deep analyses in [Younes et al 2017].

**A.1) Monocular SLAM**

**A.1.1) Filter-based SLAM.** One part of monocular SLAM is the filter-based methods. The first one is the Mono SLAM proposed by [Davison 2002] based on EKF. Later, the work is developed by them further in [Davison 2003, Davison et al 2007]. [Montemerlo and Thrun 2003] gave the monocular SLAM based on particle filter. [Strasdat et al. 2010, Strasdat et al.2012] discussed why filter-based SLAM is used that compares the filter-based methods and the keyframe-based methods. The conference paper ICRA 2010 of [Strasdat et al. 2010] achieves the best paper award, where they pointed out that keyframe-based SLAM can give more accurate results. [Nuchter et al 2007] used particle filter in SLAM to map large 3D outdoor environments. [Huang et al 2013] addressed two key limitations of the unscented Kalman filter (UKF) when applied to SLAM problem: the cubic computational complexity in the number of states and the inconsistency of the state estimates. They introduced a new sampling strategy for the UKF, which has constant computational complexity and proposed a new algorithm to ensure the unobservable subspace of the UKF's linear-regression-based system model is of the same dimension as that of the nonlinear SLAM system. [Younes et al 2017] also said that filter-based SLAM were common before 2010 and most solutions thereafter designed their systems around a non-filter, keyframe-based architecture.

**A.1.2) Keyframe-based SLAM.** The second part of monocular SLAM is the keyframe-based methods. The keyframe-based SLAM can be further categorized into: feature-based methods and direct methods. **a) Feature based SLAM:** The first keyframe-based feature SLAM is PTAM proposed by [Klein and Murray 2007]. Later the method is extended to combine edge in [Klein and Murray 2008] and extended to a mobile phone platform by them in [Klein and Murray 2009]. The key frame selections are studied by [Dong et al. 2009, Dong et al. 2014]. The SLAM++ with loop detection and object recognition is proposed by [Salas-Moreno et al. 2013]. Dynamic scene detection and adapt RANSAC is studied by [Tan et al 2013]. Specific to dynamic object, [Feng et al 2012] gave a 3D-aided optical flow SLAM. ORB SLAM of [Artal et al 2015] can deal with loop detection, dynamic scene detection, monocular, binocular, and deep images. [Bourmaud and Mégret 2015] can run in large scale environment with using submap and linear program to remove outlier. **b) Direct SLAM:** The second part of monocular SLAM is the direct methods. [Newcombe et al 2011] proposed the DTAM, the first direct SLAM, where detailed textured dense depth maps at selected keyframes are produced and meanwhile the camera's pose is tracked at frame-rate by whole image alignment against the dense textured model. A semi-dense VO is given by [Engel et al 2013]. LSD SLAM of [Engel at al 2014] provided a dense SLAM suitable to large scale environments. [Pascoe et al 2015] is a direct dense SLAM on road environments from LIDAR and cameras. A semi-dense VO on a mobile phone is performed by [Schöps et al. 2014].

### A.2) Multiocular SLAM

Multiocular SLAM uses multiple cameras to compute camera poses and 3D maps. Most of the works focus on binocular vision. They are also the bases of multiocular vision.

[Konolige and Agrawal 2008] matched visual frames with large numbers of point features using classic bundle adjustment techniques but kept only relative frame pose information. [Mei et al 2009] used local estimation of motion and structure provided by a stereo pair to represent the environment in terms of a sequence of relative locations. [Tan et al 2013] studied SLAM of multiple moving cameras in which a global map is built. [Engle et al. 2015] proposed a novel large-scale direct SLAM algorithm for stereo cameras. [Pire et al 2015] proposed a stereo SLAM system called S-PTAM which can compute the real scale of the map and overcome the limitation of PTAM when it comes to robot navigation. [Moreno et al 2016] proposed a novel approach called Sparser Relative Bundle Adjustment(SRBA) for stereo SLAM system. [Artal and Tardos 2017a] presented ORB-SLAM2 a complete SLAM system for monocular, stereo and RGB-D cameras, including map reuse, loop closing and relocalization capabilities. [Zhang et al 2015] presented a graph-based stereo SLAM system using straight lines as features. [Gomez-Ojeda et al 2017] proposed PL-SLAM, a stereo visual SLAM system that combines both points and line segments to work robustly in a wider variety of scenarios, particularly in those where point features are scarce or not well-distributed in the image. A novel direct visual-inertial odometry method for stereo cameras was proposed by [Usenko et al 2016]. [Wang et al ICCV 2017] proposed Stereo Direct Sparse Odometry (Stereo DSO) for highly accurate real-time visual odometry estimation of large-scale environments form stereo cameras. A Semi-direct Visual Odometry (SVO) for monocular and multi-camera systems was proposed by [Forster et al 2017b]. [Sun et al 2017] proposed Stereo Multi-State Constraint Kalman Filter (S-MSCKF). Compared with Multi-State Constraint Kalman Filter (MSCKF), S-MSCKF can provide significantly greater robustness.

Multiocular SLAM performs with higher reliability than monocular SLAM. In general, multiocular SLAM is preferred if hardware platforms are allowed.

**A.3) Multi-kind sensors SLAM**

Here multi-kind sensors are limited to vision and Inertial Measurement Unit (IMU). Other sensors are not introduced here. This is because recently vision and IMU fusion attracts more attention than others.

In the robotics, there are many researches of SLAM with the fusion of cameras and IMU. It is common that mobile devices are equipped with a camera and an inertial unit. Cameras can provide rich information of a scene. IMU can provide self-motion information and can also provide accurate short-term motion estimates at high frequency. Cameras and IMU have been thought to be complementary for each other. Because of universality and complementarity of visual-inertial sensors, visual-inertial fusion has been being a very active research topic in recent years. The main research approaches of visual-inertial fusion can be divided into two categories as loosely-coupled and tightly-coupled.

**A.3.1) Loosely-coupled SLAM:** In loosely-coupled systems, all sensor states are independently estimated and optimized. The integrated IMU data are incorporated as independent measurements into the stereo vision optimization in [Konolige et al 2010]. Vision-only pose estimates are used to update an extend Kalman filter (EKF) so that IMU propagation can be performed by [Weiss et al 2012]. An evaluation of different direct methods for computing frame-to-frame motion estimates of a moving sensor rig composed of an RGB-D camera and an inertial measurement unit is given and the pose from visual odometry is added to IMU optimization frame directly by [Falquez et al 2016].

**A.3.2) Tightly-coupled SLAM:** In tightly-coupled systems, all sensor states are jointly estimated and optimized. There are two approaches which are filter-based approach and nonlinear optimization-based approach for tightly-coupled systems.

**A.3.2.a) Filter-based:** The filter-based approach uses EKF to propagate and update motion states of visual-inertial sensors. MSCKF in [Mourikis and Roumeliotis 2007] uses an IMU to propagate the motion estimation of a vehicle and to update this motion estimation by observations of salient features from a monocular camera. [Li and Mourikis 2013] improved MSCKF, where a real-time EKF-based VIO algorithm, MSCKF2.0, is proposed. This algorithm can achieve consistent estimation with ensuring the correct observability properties of its linearized system model and performing online estimation of the camera-to-inertial measurement unit calibration parameters. [Li et al 2013, Li and Mourikis 2014] implemented a real-time motion tracking on a cellphone using inertial sensing and a rolling-shutter camera. MSCKF algorithm is core algorithm of Google Project Tango https://get.google.com/tango/. [Clement et al 2015] compared two modern approaches: MSCKF and Sliding Window Filter (SWF). SWF is more accurate and less sensitive to tuning parameters than MSCKF. However, MSCKF is computationally cheaper, has good consistency properties, and improves accuracies as more features are tracked. [Bloesch et al 2015] presented a monocular visual inertial odometry algorithm by directly using pixel intensity errors of image patches, where during the update step, by directly using the intensity errors as innovation term, the tracking of the multilevel patch features is closely coupled to the underlying EKF.

**A.3.2.b) Nonlinear optimization-based:** The nonlinear optimization-based approach uses keyframe-based non-linear optimization, which may potentially achieve higher accuracy due to the ability to limit linearization errors through repeated linearization of the inherently non-linear problem. [Forster et al 2017a] presented a preintegration theory that properly addresses the manifold structure of the rotation group. What's more, it is shown that the preintegration IMU model can be seamlessly integrated into a visual-inertial pipeline under the unifying framework of factor graphs. The method is short for GTSAM. [Leutenegger et al 2015] presented a novel approach which terms OKVIS to tightly integrate visual measurements with IMU measurements, where a joint non-linear cost function integrating an IMU error term with the landmark reprojection error in a fully probabilistic manner is optimized. Moreover, in order to ensure real-time operation, old states are marginalized to maintain a bounded-sized optimization window. [Li-Qin et al 2017] proposed a tightly-coupled, optimization-based, monocular visual-inertial state estimation for camera localization in complex environments. This method can run in mobile devices with a lightweight loop closure. Following ORB monocular SLAM [Artal et al 2015], a tightly-coupled visual-inertial slam system is proposed in [Artal and Tardos 2017b].

In loosely-coupled systems, it is easy to process frame data and IMU data. However, in tightly-coupled systems, in order to optimize all sensor states jointly, it is difficult to process frame data and IMU data. In terms of estimation accuracy, the tightly-coupled methods are more accurate and robust than the loosely-coupled methods. The tightly-coupled methods are more and more popular and attract more and more researches.

**B. Learning SLAM**

Learning SLAM is a new topic recently due to the deep learning development. We think it is a single category different from 3D metric SLAM and 2D topological SLAM. Learning SLAM can obtain camera pose and 3D map but need prior dataset to train a network. The performance of learning SLAM depends on the used dataset greatly and has low generalization ability. Therefore, learning SLAM is not as flexible as 3D metric SLAM and its obtained 3D map outside the used dataset is not as accurate as 3D metric SLAM at the most. But simultaneously, learning SLAM has 3D map other than 2D graph representations.

[Tateno et al 2017] used CNN to predict dense depth maps and then used keyframe-based 3D metric direct SLAM to compute camera poses. [Ummenhofer et al 2017] trained multiple stacked encoder-decoder networks to compute depth and camera motion from successive, unconstrained image pairs. [Vijayanarasimhan et al 2017] proposed a geometry-aware neural network for motion estimation in videos. [Zhou et al 2017] presented an unsupervised learning framework for estimating monocular depth and camera motion from video sequences. [Li-Wang et al 2017] proposed a monocular visual odometry system by unsupervised deep learning and by using stereo image pairs to recover the scales. [Clark et al 2017] presented an on-manifold sequence-to-sequence learning approach to motion estimation using visual and inertial sensors. [Detone et al 2017] presented a point tracking system powered by two deep convolutional neural networks MagicPoint and MagicWarp. [Gao and Zhang 2017] gave a method for loop closure detection based on the stacked denoising auto-encoder. [Araujo et al 2017] proposed a recurrent convolutional neural network based visual odometry approach for endoscopic capsule robots.

Learning SLAM increase gradually these years. However, due to lower speed and generalization ability of learning methods, in practical applications, using geometric methods is still centered.

**C. Topological SLAM**

Topological SLAM does not need accurately computing 3D map and represents environment by connectivity or topology. [Kuiper and Byun 1991] used a hierarchical description of the spatial environment, where a topological network description mediates between a control and a metrical level, also distinctive places and paths are defined by their properties at the control level, and serve as the nodes and arcs of the topological model. [Ulrich and Nourbakhsh 2000] presented a appearance-based place recognition system for topological localization. [Choset and Nagatani 2001] exploited the topology of the robot's free space to localize the robot on a partially constructed map and the topology of the environment is encoded in a generalized Voronoi graph. [ Kuipers et al 2004] described how a local perceptual map was analyzed to identify a local topology description and was abstracted to a topological place. [Chang et al 2007] presented a prediction-based SLAM algorithm to predict the structure inside an unexplored region. [Blanco et al 2008] used Bayesian filtering to provide a probabilistic estimation based on the reconstruction of the robot path in a hybrid discrete-continuous state space. [Blanco et al 2009] gave spectral graph partitioning techniques to the automatic generation of sub-maps. [Kawewong et al 2011] proposed a dictionary management to eliminate redundant searching for indoor loop-closure detection based on PIRF extraction. [Sunderhauf and Protzel 2012] presented a back-end formulation for SLAM using switchable constraints to recognize and reject outliers during loop closure detections by making the topology of the underlying factor graph representation. [Latif et al 2013] described a consensus-based approach to robust place recognition for detecting and removing past incorrect loop closures in order to deal with the problem of corrupt map estimates. [Latif et al 2014] presented a comparative analysis for graph SLAM, where graph nodes are camera poses connected by odometry or by place recognition [Vallvé et al 2018] proposed factor descent and non-cyclic factor descent, two simple algorithms for SLAM sparsification.

As shown in the above some works,topological SLAM has been developed into metric SLAM as loop detection these years. Pure studies on topological SLAM are less and less.

**D. Marker SLAM**

We have introduced the works of image based camera localization in both known and anonymous environments above. In addition, there are some works to localize cameras using some prior environment knowledge but not a 3D map such as markers. These works are regarded in semi-known environments.

In 1991, [Gatrell et al 1991] designed the concentric circular marker, which was later added additional color and scale information by [Cho et al 1998]. Ring information was considered into the marker by [Knyaz and Sibiryakov 1998]. [Kato and Billinghurst 1999] gave the first augmented reality system based on fiducial markers known as the ARToolkit, where the used marker is a black enclosed rectangle with simple graphics or texts. [Naimark and Foxlin 2002] developed a more general marker generation method, which encodes the bar code into the black circular region to produce more markers. Square marker was given by [Ababsa and Mallem 2004]. Four circles at the corners of a square was proposed by [Claus and Fitzgibbon 2005]. A black enclosed rectangle with black and white blocks was provided known as the ARTag by [Fiala 2005, Fiala 2010]. From 4 marker points, [Maidi et al 2010] developed a hybrid approach that mixes an iterative method based on the extended Kalman filter and an analytical method with direct resolution of pose parameters computation. Recently, [Bergamasco et al 2016] provided a set of circular high-contrast dots arranged into concentric layers. [DeGol et al 2017] introduced a fiducial marker ChromaTag and a detection algorithm to use opponent colors to limit and reject initial false detections and grayscale. [Munoz-Salinas et al 2018] proposed to detect keypoints for the problems of mapping and localization from a large set of squared planar markers. [Eade and Drummond 2007] gave a real-time global graph SLAM on sequences with several hundreds of landmarks. [Wu 2018] studied a new marker for camera localization without needing matching.

### 3.2.2 Structure from motion

During structure from motion procedure, camera pose computation is only a middle step. This is not the import part of image based camera localization. Therefore in the following, we give a brief introduction.

In the early stage of SFM development, there are more works on the relative pose solving. One of the useful works is the algorithm of five point relative pose of [Nister 2004], which has less degeneracies than other relative pose solvers. [Lee et al 2014] studied relative pose estimation for a multi-camera system with known vertical direction. [Kneip and Li 2014] presented a novel solution to compute the relative pose of a generalized camera. [Chatterjee and Govindu 2013] gave an efficient and robust large-scale averaging of relative 3D rotations. [Ventura et al 2015] proposed an efficient method for estimating the relative motion of a multi-camera rig from a minimal set of feature correspondences. [Fredriksson et al 2015] estimated the relative translation between two cameras and simultaneously maximizes the number of inlier correspondences.

Global pose studies are as follows. [Park et al 2014] estimate the camera direction of a geotagged image using reference images. [Carlone el at 2015] surveyed techniques for 3D rotation estimation. [Jiang et al 2013] presented a global linear method for camera pose registration. Later, the method is improved by [Cui and Tan 2015] and [Cui et al 2015].

Recently, there appear hybrid incremental and global SFM. [Cui et al 2017a, Cui et al 2017b] estimated rotations by global method and translations by incremental method and proposed community-based SFM. [Zhu et al 2017] gave parallel structure from motion from local increment to global averaging.

A recent survey on structure from motion is presented in [Ozyesil et al 2017]. Also, there are some works on learning depth from a single image. From binocular, usually disparity maps are learnt. The related works are referred to the methods ranked in the website of KITTI dataset.

### 4. Discussion

From the above techniques, we can see that currently there are less and less studies on PnP problem in small scale environment. So are studies on SFM by traditional geometric methods. But for SLAM, both traditional geometric methods and learning methods are still hot.

The studies by deep learning for image based camera localization increase gradually. However, in practical applications, using geometric methods is still centered. Deep learning methods can provide efficient image features and compensate for the geometric methods.

PnP problem or relocalization of SLAM in large scale environment still deserves studies and has not been solved well. In order for the reliability and less cost in practical applications, multi low cost sensor fusion for localization but vision sensor centered is an effective way.

In addition, some works study the pose problem of other camera sensors, such that the epipolar geometry work for a rolling shutter camera by [H. Li 2016] and radial-distorted rolling-shutter direct SLAM by [Kim et al 2017]. [Gallego et al 2017, Vidal et al 2018, Rebecq et al 2017] studied event camera SLAM.

With the hot development of SLAM, maybe it starts the age of embedding SLAM algorithm as shown by [Abouzahir et al 2018]. We think integrating the merits of all kinds of techniques is a trend for a practical SLAM system, such as geometric and learning fusion, multi-sensor fusion, multi-feature fusion, feature based ones and direct ones fusion. Integrations of these techniques may solve the current challenging difficulties such as poorly textured scenes, large illumination changes, repetitive textures, and highly dynamic motions.

## Acknowledgement

This work was supported by the National Natural Science Foundation of China under Grant No. 61421004, 61572499, 61632003.